%% file: emnlp-ijcnlp-2019.tex
\documentclass[11pt,a4paper]{article}
\usepackage[hyperref]{emnlp-ijcnlp-2019}
\usepackage{times}
\usepackage{latexsym}
\usepackage{url}

\usepackage{tabularx}
\usepackage{graphicx}
\usepackage{array}
\usepackage{multirow}
\usepackage{amsmath}
\usepackage{booktabs}
\usepackage{threeparttable}
\usepackage[toc,page]{appendix}
\usepackage{float}
\usepackage{makecell}
\usepackage{listings}
\usepackage{comment}
\usepackage{arydshln}

\aclfinalcopy %

\usepackage{xspace}
\newcommand{\bert}{\textsc{Bert}\xspace}
\newcommand{\scibert}{\textsc{SciBert}\xspace}
\newcommand{\gpt}{\textsc{GPT}\xspace}
\newcommand{\elmo}{\textsc{Elm}o\xspace}
\newcommand{\biobert}{\textsc{BioBert}\xspace}
\newcommand{\clinicalbert}{\textsc{ClinicalBert}\xspace}
\newcommand{\basevocab}{\textsc{BaseVocab}\xspace}
\newcommand{\scivocab}{\textsc{SciVocab}\xspace}

\title{\scibert: A Pretrained Language Model for Scientific Text}

\author{\makecell{Iz Beltagy~~~~~~~~Kyle Lo~~~~~~~~Arman Cohan} \\
Allen Institute for Artificial Intelligence, Seattle, WA, USA\\
{\tt $\{$beltagy,kylel,armanc$\}$@allenai.org}\\}

\date{}

\begin{document}
\maketitle
\begin{abstract}
  Obtaining large-scale annotated data for NLP tasks in the scientific domain is challenging and expensive. 
We release \scibert,
a pretrained language model based on \bert~\cite{Devlin2018BERTPO} to address the lack of high-quality, large-scale labeled scientific data. \scibert leverages unsupervised pretraining on a large multi-domain corpus of scientific publications to improve performance on downstream scientific NLP tasks.
We evaluate on a suite of tasks including
sequence tagging, sentence classification and dependency parsing, 
with datasets from a variety of scientific domains. We demonstrate statistically significant improvements over \bert and achieve new state-of-the-art results on several of these tasks. The code and pretrained models are available at 
\url{https://github.com/allenai/scibert/}.
\end{abstract}

\input{1-intro}

\input{2-method.tex}

\input{3-experiments.tex}

\input{both-tables_without_bio.tex}

\input{4-results.tex}

\input{5-discussion.tex}

\input{6-conclusion.tex}

\bibliography{emnlp-ijcnlp-2019}
\bibliographystyle{acl_natbib}

\end{document}

%% file: 1-intro.tex
\section{Introduction}
The exponential increase in the volume of scientific publications in the past decades has made NLP an essential tool for large-scale knowledge extraction and machine reading of these documents. 
Recent progress in NLP has been driven by the adoption of deep neural models, but training such models often requires large amounts of labeled data. 
In general domains, large-scale training data is often possible to obtain through crowdsourcing, but in scientific domains, annotated data is difficult and expensive to collect due to the expertise required for quality annotation.

As shown through \elmo~\cite{Peters2018DeepCW}, \gpt~\cite{radford2018improving} and \bert~\cite{Devlin2018BERTPO},
unsupervised pretraining of language models on large corpora significantly improves performance on many NLP tasks.
These models return contextualized embeddings for each token which can be passed into minimal task-specific neural architectures.
Leveraging the success of unsupervised pretraining has become especially important especially when task-specific annotations are difficult to obtain, like in scientific NLP.  Yet while both \bert and \elmo have released pretrained models, they are still trained on general domain corpora such as news articles and Wikipedia.  

In this work, we make the following contributions:

\indent (\textit{i}) We release \scibert, a new resource demonstrated to improve performance on a range of NLP tasks in the scientific domain. \scibert is a pretrained language model based on \bert but trained on a large corpus of scientific text.

\indent (\textit{ii}) We perform extensive experimentation to investigate the performance of finetuning versus task-specific architectures atop frozen embeddings, and the effect of having an in-domain vocabulary.

\indent (\textit{iii}) We evaluate \scibert on a suite of tasks in the scientific domain, and achieve new state-of-the-art (SOTA) results on many of these tasks.

%% file: 2-method.tex
\section{Methods}

\paragraph{Background} The \bert model architecture \cite{Devlin2018BERTPO} is based on a multilayer bidirectional Transformer \cite{Vaswani2017AttentionIA}. Instead of the traditional left-to-right language modeling objective, \bert is trained on two tasks: predicting randomly masked tokens and predicting whether two sentences follow each other. \scibert follows the same architecture as \bert but is instead pretrained on scientific text.

\paragraph{Vocabulary}
\label{sec:vocab}
\bert uses WordPiece~\cite{ Wu2016GooglesNM} for unsupervised tokenization
of the input text. 
The vocabulary is built such that it contains
the most frequently used words or subword units.  %
We refer to the original vocabulary released with \bert as \basevocab.

We construct \scivocab, a new WordPiece vocabulary on our scientific corpus using the SentencePiece\footnote{\url{https://github.com/google/sentencepiece}}
library.  
We produce both cased and uncased vocabularies and set the vocabulary size to 30K to match the size of \basevocab. The resulting token overlap between \basevocab and \scivocab is 42\%, illustrating a substantial difference in frequently used words between scientific and general domain texts.

\paragraph{Corpus}
We train \scibert on a random sample of 1.14M papers from
Semantic Scholar~\cite{ammar:18}. 
This corpus consists of 18\% papers from 
the computer science domain and 82\% from the broad biomedical domain. 
We use the full text of the papers, not just the abstracts.
The average paper length is 154 sentences (2,769 tokens) 
resulting in a corpus size of 3.17B tokens, similar to the 3.3B tokens on which \bert was trained. We split sentences using ScispaCy~\cite{Neumann2019},\footnote{\url{https://github.com/allenai/SciSpaCy}} which is optimized for scientific text.

%% file: 3-experiments.tex
\section{Experimental Setup}

\subsection{Tasks}

We experiment on the following core NLP tasks: 
\begin{enumerate}
  \setlength{\itemsep}{0pt}
  \setlength{\parskip}{0pt}
    \item Named Entity Recognition (NER)
    \item PICO Extraction (PICO)
    \item Text Classification (CLS)
    \item Relation Classification (REL)
    \item Dependency Parsing (DEP)
\end{enumerate}
PICO, like NER, is a sequence labeling task where the model extracts spans describing the Participants, Interventions, Comparisons, and Outcomes in a clinical trial paper \cite{Kim2011AutomaticCO}.  REL is a special case of text classification where the model predicts the type of relation expressed between two entities, which are encapsulated in the sentence by inserted special tokens.

\subsection{Datasets}

For brevity, we only describe the newer datasets here, and refer the reader to the references in Table~\ref{tab:results} for the older datasets. 
EBM-NLP \cite{Nye2018ACW} annotates PICO spans in clinical trial abstracts. 
SciERC \cite{Luan2018MultiTaskIO} annotates entities and relations from computer science abstracts.  
ACL-ARC \cite{Jurgens2018MeasuringTE} and SciCite \cite{naacl2019-scicite} assign intent labels (e.g. Comparison, Extension, etc.) to sentences from scientific papers that cite other papers.  
The Paper Field dataset is built from the Microsoft Academic Graph~\cite{Sinha2015AnOO}\footnote{\url{https://academic.microsoft.com/}} and maps paper titles to one of 7 fields of study.
Each field of study (i.e. geography, politics, economics, business, sociology, medicine, and psychology) has approximately 12K training examples.

\subsection{Pretrained \bert Variants}

\paragraph{\bert-Base} We use the pretrained weights for \bert-Base \cite{Devlin2018BERTPO} released with the original \bert code.\footnote{\url{https://github.com/google-research/bert}} The vocabulary is \basevocab. We evaluate both cased and uncased versions of this model.

\paragraph{\scibert}
We use the original \bert code
to train \scibert on our corpus with the same configuration and size as \bert-Base. 
We train 4 different versions of \scibert:
(\textit{i}) cased or uncased and (\textit{ii}) \basevocab or \scivocab. 
The two models that use \basevocab are finetuned from the corresponding \bert-Base models. 
The other two models that use the new \scivocab are 
trained from scratch.

Pretraining \bert for long sentences can be slow.  Following the original \bert code, we set a maximum sentence length of 128 tokens, and train the model until the training loss stops decreasing. We then continue training the model allowing sentence lengths up to 512 tokens. 

We use a single TPU v3 with 8 cores.  Training the \scivocab models from scratch on our corpus takes 1 week\footnote{\bert's largest model was trained on 16 Cloud TPUs for 4 days. Expected 40-70 days~\cite{timdettmers} on an 8-GPU machine.} (5 days with max length 128, then 2 days with max length 512). 
The \basevocab models take 2 fewer days of training because they aren't trained from scratch.

All pretrained \bert models are converted to be compatible with PyTorch using the pytorch-transformers library.\footnote{\url{https://github.com/huggingface/pytorch-transformers}}  All our models (Sections~\ref{sec:finetune} and \ref{sec:frozen}) are implemented in PyTorch using AllenNLP~\cite{Gardner2017AllenNLP}.  

\paragraph{Casing}
We follow \citet{Devlin2018BERTPO} in using the cased models for NER and the uncased models for all other tasks.  We also use the cased models for parsing.  Some light experimentation showed that the uncased models perform slightly better (even sometimes on NER) than cased models.

\subsection{Finetuning \bert}
\label{sec:finetune}

We mostly follow the same architecture, optimization, and hyperparameter choices used in \citet{Devlin2018BERTPO}.  For text classification (i.e. CLS and REL), we feed the final \bert vector for the \texttt{[CLS]} token into a linear classification layer.  For sequence labeling (i.e. NER and PICO), we feed the final \bert vector for each token into a linear classification layer with softmax output.  We differ slightly in using an additional conditional random field, which made evaluation easier by guaranteeing well-formed entities.  For DEP, we use the model from \citet{Dozat2017DeepBA} with dependency tag and arc embeddings of size 100 and biaffine matrix attention over \bert vectors instead of stacked BiLSTMs.

In all settings, we apply a dropout of 0.1 and optimize cross entropy loss using Adam \cite{Kingma2015AdamAM}.  We finetune for 2 to 5 epochs using a batch size of 32 and a learning rate of 5e-6, 1e-5, 2e-5, or 5e-5 with a slanted triangular schedule \cite{Howard2018UniversalLM} which is equivalent to the linear warmup followed by linear decay~\cite{Devlin2018BERTPO}.  For each dataset and \bert variant, we pick the best learning rate and number of epochs on the development set and report the corresponding test results.

We found the setting that works best across most datasets and models is 2 or 4 epochs and a learning rate of 2e-5.  While task-dependent, optimal hyperparameters for each task are often the same across \bert variants.

\subsection{Frozen \bert Embeddings}
\label{sec:frozen}

We also explore the usage of \bert as pretrained contextualized word embeddings, like ELMo \cite{Peters2018DeepCW}, by training simple task-specific models atop frozen \bert embeddings.  

For text classification, we feed each sentence of \bert vectors into a 2-layer BiLSTM of size 200 and apply a multilayer perceptron (with hidden size 200) on the concatenated first and last BiLSTM vectors.  For sequence labeling, we use the same BiLSTM layers and use a conditional random field to guarantee well-formed predictions.  For DEP, we use the full model from \citet{Dozat2017DeepBA} with dependency tag and arc embeddings of size 100 and the same BiLSTM setup as other tasks.  We did not find changing the depth or size of the BiLSTMs to significantly impact results \cite{Reimers2017OptimalHF}.  

We optimize cross entropy loss using Adam, but holding \bert weights frozen and applying a dropout of 0.5.  We train with early stopping on the development set (patience of 10) using a batch size of 32 and a learning rate of 0.001.

We did not perform extensive hyperparameter search, but while optimal hyperparameters are going to be task-dependent, some light experimentation showed these settings work fairly well across most tasks and \bert variants.

%% file: both-tables_without_bio.tex
\setlength{\dashlinedash}{0.2pt}
\setlength{\dashlinegap}{1.5pt}
\setlength{\arrayrulewidth}{0.4pt}

\begin{table*}[ht]
\centering
\small
\setlength\tabcolsep{8pt}
\renewcommand{\arraystretch}{1.1}
\begin{tabular}{@{}lllccccr@{}}
\toprule
Field & Task & Dataset & SOTA & \multicolumn{2}{c}{\bert-Base} & \multicolumn{2}{c}{\scibert} \\
\cdashline{5-8}
& & & &  Frozen & Finetune & Frozen & Finetune \\
\midrule

\multirow{5}{*}{Bio} & \multirow{3}{*}{NER}
& BC5CDR \cite{Li2016BioCreativeVC} & 88.85\footnotemark & 85.08 & 86.72 & 88.73 & \textbf{90.01}  \\
&& JNLPBA \cite{Collier2004IntroductionTT}  & \textbf{78.58} & 74.05 & 76.09 & 75.77 & 77.28 \\
&& NCBI-disease ~\cite{Dogan2014NCBIDC}   & \textbf{89.36} & 84.06 & 86.88 & 86.39 & 88.57 \\
\cdashline{2-8}
 & PICO  & EBM-NLP \cite{Nye2018ACW} & 66.30 & 61.44 & 71.53 & 68.30 & \textbf{72.28}  \\
\cdashline{2-8}
& \multirow{2}{*}{DEP}
 & GENIA~\cite{Kim2003GENIAC} - LAS &  \textbf{91.92}  & 90.22 & 90.33 & 90.36  & 90.43 \\
 & & GENIA~\cite{Kim2003GENIAC} - UAS & \textbf{92.84}  & 91.84 & 91.89 & 92.00 & 91.99 \\
\cdashline{2-8}
& REL
& ChemProt \cite{Kringelum2016ChemProt30AG} & 76.68 & 68.21 & 79.14 & 75.03 & \textbf{83.64} \\

\midrule

\multirow{3}{*}{CS} & \multirow{1}{*}{NER}
& SciERC \cite{Luan2018MultiTaskIO} & 64.20 & 63.58 & 65.24 &  65.77 & \textbf{67.57} \\
\cdashline{2-8}
& \multirow{1}{*}{REL}
& SciERC  \cite{Luan2018MultiTaskIO}     & n/a & 72.74 & 78.71 & 75.25 & \textbf{79.97} \\
\cdashline{2-8}
& CLS
& ACL-ARC \cite{Jurgens2018MeasuringTE} & 67.9 & 62.04 & 63.91 &  60.74 & \textbf{70.98} \\

\midrule

\multirow{2}{*}{Multi} & \multirow{2}{*}{CLS}
& Paper Field  & n/a & 63.64 & 65.37 & 64.38 & \textbf{65.71} \\
&& SciCite \cite{naacl2019-scicite}  & 84.0 & 84.31 & 84.85 & \textbf{85.42} & \textbf{85.49} \\

\midrule

Average &  &  & & 73.58 & 77.16 & 76.01 & 79.27  \\

\bottomrule
\end{tabular}
\caption{
Test performances of all \bert variants on all tasks and datasets. 
\textbf{Bold} indicates the SOTA result (multiple results bolded if difference within 95\% bootstrap confidence interval).
Keeping with past work, we report macro F1 scores for NER (span-level), macro F1 scores for REL and CLS (sentence-level), and macro F1 for PICO (token-level), and micro F1 for ChemProt specifically.
For DEP, we report labeled (LAS) and unlabeled (UAS) attachment scores (excluding punctuation) for the same model with hyperparameters tuned for LAS.
All results are the average of multiple runs with different random seeds.
}
\label{tab:results}
\end{table*}
\footnotetext{The SOTA paper did not report a single score.  We compute the average of the reported results for each class weighted by number of examples in each class.}

%% file: 4-results.tex
\section{Results}
\label{sec:results}
Table~\ref{tab:results} summarizes the experimental results. We observe that \scibert outperforms \bert-Base on scientific tasks (+2.11 F1 with finetuning and +2.43 F1 without)\footnote{For rest of this paper, all results reported in this manner are averaged over datasets excluding UAS for DEP since we already include LAS.}.  We also achieve new SOTA results on many of these tasks using \scibert.

\subsection{Biomedical Domain} 
We observe that \scibert outperforms \bert-Base on biomedical tasks (+1.92 F1 with finetuning and +3.59 F1 without). In addition, \scibert achieves new SOTA results on BC5CDR and ChemProt~\cite{Lee2019BioBERTAP}, and EBM-NLP~\cite{Nye2018ACW}.

\scibert performs slightly worse than SOTA on 3 datasets.  The SOTA model for JNLPBA is a BiLSTM-CRF ensemble trained on multiple NER datasets not just JNLPBA~\cite{Yoon2018CollaboNetCO}.  The SOTA model for NCBI-disease is \biobert~\cite{Lee2019BioBERTAP}, which is \bert-Base finetuned on 18B tokens from biomedical papers.  The SOTA result for GENIA is in ~\citet{Nguyen2019FromPT} which uses the model from \citet{Dozat2017DeepBA} with part-of-speech (POS) features, which we do not use.

In Table~\ref{tab:biobert}, we compare \scibert results with reported \biobert results on the subset of datasets included in \cite{Lee2019BioBERTAP}.  Interesting, \scibert outperforms \biobert results on BC5CDR and ChemProt, and performs similarly on JNLPBA despite being trained on a substantially smaller biomedical corpus.

\begin{table}[tb]
\centering
\small
\setlength\tabcolsep{8pt}
\renewcommand{\arraystretch}{1.1}
\begin{tabular}{@{}llrr@{}}
\toprule
Task & Dataset & \biobert & \scibert\\
\midrule
\multirow{3}{*}{NER}
& BC5CDR & 88.85 & 90.01  \\
& JNLPBA  & 77.59 & 77.28 \\
& NCBI-disease  & 89.36 & 88.57 \\
\cdashline{2-4}
REL & ChemProt & 76.68 & 83.64 \\
\bottomrule
\end{tabular}
\caption{Comparing \scibert with the reported \biobert results on biomedical datasets.}
\label{tab:biobert}
\end{table}

\subsection{Computer Science Domain}
We observe that \scibert outperforms \bert-Base on computer science tasks (+3.55 F1 with finetuning and +1.13 F1 without). In addition, \scibert achieves new SOTA results on ACL-ARC~\cite{naacl2019-scicite}, 
and the NER part of SciERC~\cite{Luan2018MultiTaskIO}. 
For relations in SciERC, our results are not comparable with those in \citet{Luan2018MultiTaskIO} because we are performing relation classification given gold entities, while they perform joint entity and relation extraction. 

\subsection{Multiple Domains}
We observe that \scibert outperforms \bert-Base on the multidomain tasks (+0.49 F1 with finetuning and +0.93 F1 without).  In addition, \scibert outperforms the SOTA on SciCite~\cite{naacl2019-scicite}.  No prior published SOTA results exist for the Paper Field dataset.

%% file: 5-discussion.tex
\section{Discussion}

\subsection{Effect of Finetuning}

We observe improved results via \bert finetuning rather than task-specific architectures atop frozen embeddings (+3.25 F1 with \scibert and +3.58 with \bert-Base, on average).  For each scientific domain, we observe the largest effects of finetuning on the computer science (+5.59 F1 with \scibert and +3.17 F1 with \bert-Base) and biomedical tasks (+2.94 F1 with \scibert and +4.61 F1 with \bert-Base), and the smallest effect on multidomain tasks (+0.7 F1 with \scibert and +1.14 F1 with \bert-Base).  On every dataset except BC5CDR and SciCite, \bert-Base with finetuning outperforms (or performs similarly to) a model using frozen \scibert embeddings.

\subsection{Effect of \scivocab}

We assess the importance of an in-domain scientific vocabulary by repeating the finetuning experiments for \scibert with \basevocab.  We find the optimal hyperparameters for \scibert-\basevocab often coincide with those of \scibert-\scivocab.  

Averaged across datasets, we observe +0.60 F1 when using \scivocab.  For each scientific domain, we observe +0.76 F1 for biomedical tasks, +0.61 F1 for computer science tasks, and +0.11 F1 for multidomain tasks.  

Given the disjoint vocabularies (Section~\ref{sec:vocab}) and the magnitude of improvement over \bert-Base (Section~\ref{sec:results}), we suspect that while an in-domain vocabulary is helpful, \scibert benefits most from the scientific corpus pretraining.

%% file: 6-conclusion.tex
\section{Related Work}
Recent work on domain adaptation of BERT includes 
\biobert~\cite{Lee2019BioBERTAP} and \clinicalbert~\cite{Alsentzer2019PubliclyAC,clinicalbert}.
\biobert is trained on PubMed abstracts and PMC full text articles, and \clinicalbert is trained on clinical text from the MIMIC-III database~\cite{mimic}.
In contrast, \scibert is trained on the full text of 1.14M biomedical and computer science papers from the Semantic Scholar corpus \cite{ammar:18}. 
Furthermore, \scibert uses an in-domain vocabulary (\scivocab) while the other above-mentioned models use the original \bert vocabulary (\basevocab). 

\section{Conclusion and Future Work}
We released \scibert, a pretrained language model for scientific text based on \bert.
We evaluated \scibert on a suite of tasks and datasets from scientific domains. \scibert significantly outperformed \bert-Base and achieves new SOTA results on several of these tasks, even compared to some reported \biobert~\cite{Lee2019BioBERTAP} results on biomedical tasks.

For future work, we will release a version of \scibert analogous to \bert-Large, as well as experiment with different proportions of papers from each domain.  Because these language models are costly to train, we aim to build a single resource that's useful across multiple domains.

\section*{Acknowledgment}
We thank the anonymous reviewers for their comments and suggestions.  We also thank Waleed Ammar, Noah Smith, Yoav Goldberg, Daniel King, Doug Downey, and Dan Weld for their helpful discussions and feedback.
All experiments were performed on \url{beaker.org} and supported in part by credits from Google Cloud.